\pdfoutput=1

\documentclass[11pt]{article}

\usepackage{ACL2023}

\usepackage{times}
\usepackage{latexsym}
\usepackage{graphics}
\usepackage{url}

\usepackage[whole]{bxcjkjatype}
\usepackage{multirow}

\usepackage[T1]{fontenc}

\usepackage[utf8]{inputenc}

\usepackage{microtype}

\usepackage{inconsolata}

\newcommand{\syllogism}[3]{
\begin{tabular}{l}
\textbf{P1}: #1\\
\textbf{P2}: #2\\ \hline
\textbf{C}: #3
\end{tabular}
}

\newcommand{\ShowExp}[7]{
\multirow{6}{*}{#1} &
\multirow{3}{*}{English} & \textbf{P1}: #2\\ & & \textbf{P2}: #3\\ & & \textbf{C}: #4\\ \cline{2-3}
& \multirow{3}{*}{Japanese} & \textbf{P1}: #5\\ & & \textbf{P2}: #6\\ & & \textbf{C}: #7
}

\title{Evaluating Large Language Models with NeuBAROCO:\\
Syllogistic Reasoning Ability and Human-like Biases}

\author{Risako Ando \\
  \texttt{risakochaan@keio.jp} \\\And
  Takanobu Morishita \\
  \texttt{morishita@keio.jp} \\\And
  Hirohiko Abe \\
  \texttt{hirohiko-abe@keio.jp} \\
  \AND
  Koji Mineshima\\
  \texttt{minesima@abelard.flet.keio.ac.jp} \\\And
  Mitsuhiro Okada\\
  \texttt{mitsu@abelard.flet.keio.ac.jp} \\
  \AND
  Keio University 
  }

\begin{document}
\maketitle

\begin{abstract}
This paper investigates whether current large language models exhibit biases in logical reasoning, similar to humans. Specifically, we focus on syllogistic reasoning, a well-studied form of inference in the cognitive science of human deduction. To facilitate our analysis, we introduce a dataset called NeuBAROCO, originally designed for psychological experiments that assess human logical abilities in syllogistic reasoning.
The dataset consists of syllogistic inferences in both English and Japanese.
We examine three types of biases observed in human syllogistic reasoning: belief biases, conversion errors, and atmosphere effects. Our findings demonstrate that current large language models struggle more with problems involving these three types of biases.

\end{abstract}

\section{Introduction}
\label{sec:intro}

Syllogistic inferences and their various variants have been extensively studied since \textit{Prior Analytics} by Aristotle in the 4th century BC.
While the Aristotelean syllogism is a small part of the predicate logic and a limited inference system when compared to, for example, the formal system of logical inference rules, there has been recently a revival movement of Aristotelean syllogism and its variants, including natural logic~\cite{vanBenthem86,Sanchez91,moss2015natural}.
This renewed attention arises from the perspective of viewing syllogistic inferences as a ``natural'' inference rule applicable to our everyday reasoning in ordinary language.

Not only is there a re-evaluation of the significance of syllogistic inferences and their variants in relation to their usefulness in ordinary language, but they are also considered as a benchmark for various inference studies in different disciplines.
For example, cognitive psychological studies of logical inferences~\cite{stenning2012human}, diagrammatic logical inference studies~\cite{sato2015diagrams},
neuroscientific studies of logical inferences~\cite{goel2000dissociation},
all draw upon syllogistic reasoning as a point of reference.
On the other hand, the recent developments of deep-learning-based AI-tools of natural languages,
in particular, the state-of-the art Large Language Models (LLMs), including BERT~\cite{devlin-etal-2019-bert}
and GPT~\cite{brown2020language}, are remarkable. These tools hold the potential to be useful for logical inferences.
However, there is still a need for further accumulation of studies on the use of AI models for logical inferences in natural language.

In this paper, we explore the potential of LLMs for performing logical inferences, with a specific focus on using syllogistic inferences as a benchmark.
We present the NeuBAROCO dataset, a new dataset consisting of syllogistic inferences in both English and Japanese.
The dataset is derived from the question collection BAROCO~\cite{SHIKISHIMA2009256}, which has been used in various studies in Japan to evaluate human syllogistic reasoning abilities.

The field of cognitive science of human reasoning has not yet been fully integrated with the recent advancements in AI, despite its potential to provide valuable insights for AI inference research. Considering the significant attention given to biases in studies of logical inference within cognitive psychology and cognitive science~\cite{evans1989bias,evans1993human}, our primary focus lies in assessing whether LLMs exhibit the biases observed in human logical inferences.
We focus on three types of human biases that have been studied in cognitive science, namely, belief biases, conversion errors, and atmosphere effects (see Section \ref{sec:bias}).
We explore the extent to which currently available LLMs for natural language inference can effectively address syllogistic inferences that are susceptible to errors resulting from these biases.

The contributions of the paper are summarized as follows.

\begin{enumerate}
\item We present the NeuBAROCO dataset, specifically designed for syllogistic inferences, which serves as a valuable resource for examining human-biases in language models.
\item Using this dataset, we evaluate the logical reasoning ability of several recent LLMs both for English and Japanese.
\item Our evaluation reveals that the current LLMs exhibit significant shortcomings when faced with problems that are prone to errors resulting from the three biases of interest.
\end{enumerate}

The dataset will be made publicly available for research on human and machine understanding of logical inferences.

\section{Background and related Work}
\label{sec:background}

\subsection{Syllogism}
\label{ssec:logic}

A syllogism consists of the following 
four types of categorical sentences 
described as \textbf{A}, \textbf{E}, \textbf{I}, and \textbf{O}.

\begin{table}[h]
\centering
\scalebox{0.9}{
\begin{tabular}{cll}  \hline
Type & Form & Description \\ \hline
\textbf{A} & {\sf All $S$ are $P$} & Universal affirmative \\
\textbf{E} & {\sf No $S$ are $P$} & Universal negative \\
\textbf{I} & {\sf Some $S$ are $P$} & Particular affirmative \\
\textbf{O} & {\sf Some $S$ are not $P$} & Particular negative \\ \hline
\end{tabular}
}
\caption{Four types of categorical sentences}
\label{tab:categorical}
\end{table}
Each syllogism is composed of
two premises (\textbf{P1}, \textbf{P2}) and one conclusion (\textbf{C}).
Thus, each type of syllogisms is identified by the types of three sentences.
The following is an example of EIO-type syllogism (called \textit{Ferio}
in the traditional mnemonic), which is a valid pattern of syllogism.

\begin{center}
\syllogism{No B are C. (\textbf{E}-type)}
{Some A are B. (\textbf{I}-type)}
{Some A are not C. (\textbf{O}-type)}
\end{center}

As outlined in Section \ref{sec:intro},
syllogisms have been extensively studied in the fields of logic and psychology.
To assess the logical reasoning capabilities of current language models in natural language inferences, we use a dataset that encompasses various types of syllogistic inferences.

\subsection{Machine learning and logical inference}
\label{ssec:ML}

In recent years, syllogistic inferences and their variants have been used to examine the reasoning ability of machine learning based models for natural language inferences (NLI).

\citet{richardson2020probing} examined the capacities of NLI models to perform various types of logical inferences
that involve boolean operators, quantifiers, comparatives, conditionals, negation, and counting expressions. The study utilized synthetically generated data. Their findings indicate that models fine-tuned with NLI datasets perform well, suggesting that NLI models enhance their accuracy when provided with additional datasets as input.
However, due to the artificial nature of the data, the majority of the inferences in each type of inferences
share a similar structure,
making it relatively easy for machine learning algorithms to solve such similar problems.
In contrast, our study utilizes human-generated data, with a specific emphasis on inferences that elicit human-like biases. Furthermore, our study focuses on the capacity of LLMs to solve syllogistic inferences without requiring fine-tuning with NLI training examples.

\citet{yanaka2019can} examined monotoniticy inferences, those logical inferences that are licensed by substituting general terms in quantified sentences.
A testset for monotonicity inference was created semi-automatically and then tested on various language models.
One interesting finding is that all the models performed poorly on a class of inferences involving negative contexts (the so-called downward monotone inferences).
Monotonicity inferences are structurally simpler than syllogistic inferences in that they consist of single-premises; by contrast, syllogisms are composed of multiple premises. Combination of quantifiers in syllogisms such as \textit{no} and \textit{some} can be more challenging than monotonicity inferences.

\citet{schlegel-etal-2022-transformers} conducted an empirical study to investigate the detection of formally valid inference within controlled fragments of natural language. These fragments were specifically designed to increase the complexity of the satisfiability problem. In their study, each fragment consisted of 
artifically generated sets of English sentences that incorporated determiners such as \textit{every}, \textit{some}, \textit{no}, negation \textit{not}, and relative clauses, within the context
of a vocabulary comprising count nouns and transitive verbs.
The findings of \citet{schlegel-etal-2022-transformers} indicate that transformer-based language models 
fine-tuned with training data
tend to exhibit overfitting to superficial patterns present in the data, rather than acquiring the logical principles that govern reasoning within these fragments. In other words, the models seem to focus on surface-level features rather than grasping the underlying logical principles.
Furthermore, according to \citet{schlegel-etal-2022-transformers},
the ability of neural networks to learn and solve the various satisfiability problems does not appear to align with the complexity classes associated with the elicited fragments.
Our study focuses on a small yet manually controlled dataset, as opposed to a large corpus of artificially generated data. Specifically, we manually annotate problems that are susceptible to human-like biases with corresponding labels. This approach facilitates meaningful comparisons between the syllogistic inference capabilities of humans and models.

Closest to our work is \citet{dasgupta2022language},
which reveals that large language models show content effects
(i.e., what we called \textit{belief biases}) in syllogism reasoning as well as humans.
They introduced new datasets of abstract logical inferences including syllogisms.
Each syllogism is annotated with
the information about whether or not a proposition is consistent with human beliefs and knowledge.
They found that when the conclusion of an inference contradicts reality, the language model exhibits a strong bias towards classifying the argument as invalid, regardless of its logical validity.
Our experimental results provide further support of these findings and demonstrate that similar effects are observed
not only in English but also in Japanese, 
a typologically different language from English.
Furthermore, we expand our focus beyond belief biases to include various types of biases such as conversion errors and atmosphere effects. We systematically examine the impact of these biases on LLMs with the ultimate objective of comparing the performance of LLMs in logical reasoning to that of humans.

\section{The NeuBAROCO dataset}
\label{sec:dataset}

\subsection{Background: the BAROCO dataset}
\label{ssec:orig}

BAROCO is the collection of logical inference questions to examine subjects' ability of logical inference. The questions of BAROCO are mainly composed of syllogistic inferences and their variants. BAROCO has been used in various studies on human logical inference abilities. BAROCO was first used for behavioral genetic studies with the twin method in \citet{shikishima2006be,SHIKISHIMA2009256},
where the genetic factor and the environmental factor of the logical inference ability were measured. 500 twin pairs (1,000 participants) were asked to answer 100 questions about a version of BAROCO. 
The results were then compared with the subjects' scores on a standard IQ test that typically did not include logical inference abilities.
Additionally, correlations were explored between logical inference abilities and decision-making skills in the fields of behavioral economics and cognitive sociology. 
For instance, \citet{shikishima2015genetic} investigated the relationship between logical inference abilities and Allais's decision-making task, along with other related studies.


\subsection{Data construction}
\label{dataconst}

The full version of the original BAROCO dataset comprises a collection of 209 logical inferences divided into seven sections, each containing different types of questions. The version of BAROCO called ``BAROCO-ALL'' encompasses a total of 200 questions, which includes the following three sections. Examples of each section will be presented in Table~\ref{tab:examples-symcon}.
\begin{itemize}
\item[(1)] Abstract syllogism inferences: This section consists of inferences where the terms used in the sentences are represented by capital letters of the alphabet.
\item[(2)] Contentual (belief-consistent) syllogistic inferences: In this section, the inferences are constructed using concrete nouns commonly used in ordinary language.
\item[(3)] Belief-inconsistent syllogistic inferences: This section introduces inferences where belief-inconsistent sentences may appear within the inference itself.
\end{itemize}

Most questions in BAROCO dataset consist of two premises and three options for the correct answer.
In the original setup, the participants were asked to choose one logically valid conclusion from the given options.
We transformed each question into a format commonly used for evaluating NLI models, where inferences are categorized as \textit{entailment}, \textit{contradiction}, or \textit{neither} (which we call \textit{neutral}).
In accordance with the format of syllogisms, each inference consists of two premises and one conclusion.
We manually assigned each inference with the approapriate label of \textit{entailment}, \textit{contradiction}, and \textit{neutral}.

The resulting dataset obtained from this process is referred to as NeuBAROCO.
In total, there are 375 inference problems in the NeuBAROCO dataset, with 122 instances labelled as \textit{entailment},
71 instances labelled as \textit{contradiction},
and 182 instances labelled as \textit{neutral}.

\begin{table*}[ht!]
\centering
\begin{tabular}{l | l | l} \hline
\textbf{Type} & \textbf{Language} & \textbf{Example} \\ \hline
\ShowExp{Symbol}
{All A are B.}
{All B are C.}
{All A are C.}
{すべての \ A \ は \ B \ である}
{すべての \ B \ は \ C \ である}
{すべての \ A \ は \ C \ である}
\\ \hline\hline
\ShowExp{Consistent}
{One friend of Taro is a friend of Paul.}
{All of Paul's friends are German.}
{One of Taro's friends is German.}
{太郎のある友人はポールの友人である。}
{ポールのすべての友人はドイツ人である。}
{太郎のある友人はドイツ人である。}
\\ \hline\hline
\ShowExp{Inconsistent}
{Some animals are human beings.}
{All animals are tomatoes.}
{Some humans are tomatoes.}
{ある動物は人間である。}
{すべての動物はトマトである。}
{ある人間はトマトである。}

\\ \hline
\end{tabular}
\caption{Examples of symbolic, consistent, and inconsistent 
syllogism in the NeuBAROCO dataset.
The English sentences (\textbf{P1}, \textbf{P2}, \textbf{C})
in each example correspond to the respective Japanese sentences.
The correct label for all examples is \textit{entailment}.
}
\label{tab:examples-symcon}
\end{table*}

The BAROCO dataset was written in Japanese. We translated each problem into English using the DeepL translation tool
(\url{https://www.deepl.com/translator}). We manually checked and adjusted the wording of each sentence,
ensuring that they conform to the patterns of categorical sentences.
We normalized the quantifiers in the English sentences.
We used \textit{all} or \textit{every} for universal quantification in \textbf{A}-type sentences, and \textit{some}, \textit{a certain}, or \textit{one of} for existential quantification in \textbf{I}-type and \textbf{O}-type sentences,
and \textit{no} for universal negative in \textbf{E}-type sentences.
To prevent the sentences from being interpreted as generic statements, we refrained from using the indefinite article \textit{a} (or \textit{an}) for existential quantification. The presence of the indefinite article can lead to a generic interpretation, such as in the sentence \textit{A cat is an animal}.
This ensures consistency and clarity in the translation of the original Japanese sentences into English.

\subsection{Annotation}
\label{ssec:overview}

We annotated each inference problem in the NeuBAROCO dataset
as to what type of inference it is and whether
the sentences appearing in it are consistent with beliefs.

\subsubsection{Types of logical inferences}

There are two types of inferences in the dataset:
basic syllogisms and extended syllogisms.

\paragraph{Basic syllogisms}
As explained in Section \ref{sec:background},
basic syllogisms consist of two premises (\textbf{P1}, \textbf{P2})
and one conclusion (\textbf{C}).
We annotate each basic syllogism
with the types of premises and conclusion.
The following is an example of \textbf{IAI}-type syllogism:
\begin{center}
\syllogism{Some A are B. (\textbf{I}-type)}
{All B are C. (\textbf{A}-type)}
{Some A are C. (\textbf{I}-type)}
\end{center}
The first premise is \textbf{I}-type. The second premise is \textbf{A}-type. The conclusion is \textbf{I}-type. Therefore, the inference is labeled as \textbf{IAI}. 

\paragraph{Extended syllogisms}
Extended syllogisms can be classified into two types.
One is a boolean inference where conjunction \textit{and}
and \textit{or} appear between terms.
The following is an example:
\begin{center}
\syllogism{All A or B are C.}
{No C are D.}
{No B are D.}
\end{center}

The other is a hypothetical syllogism,
one of whose premises is a conditional sentence
of the form \textit{If P then Q}.
Here, \textit{P} or \textit{Q} can be a negated sentence.
The following is an example:

\begin{center}
\scalebox{0.9}{
\syllogism{If Hanako has blood type O,\\
then Hanako's daughter has blood type B.}
{Hanako's daughter does not have blood type B.}
{Hanako does not have blood type O.}
}
\end{center}

In the dataset,
there are 318 basic syllogisms
and 57 extended syllogisms.

\subsubsection{Belief consistency}
\label{sssec:belief}

We also classify the inferences into three distinct types based on the types of sentences they contain: \textit{symbolic}, \textit{consistent}, and \textit{inconsistent}.
Table \ref{tab:examples-symcon} shows examples of each type.

\paragraph{Symbolic} A symbolic inference is composed of sentences where
all the terms are abstract symbols (alphabets).
For humans, they can be considered to be neutral with respect to beliefs.

\paragraph{Consistent} An inference is labelled as \textit{consistent} if all of the premises and conclusion
are consistent with common-sense beliefs. 
In the case of the example in Table \ref{tab:examples-symcon},
all the sentences, i.e.,
\textit{One friend of Taro is a friend of Paul},
\textit{All of Paul's friends are German},
and \textit{One of Taro's friends is German},
can be interpreted consistent with belief.

\paragraph{Inconsistent}
An inference is labelled as \textit{inconsistent} if at least one of of the premises and conclusion is inconsistent with common-sense beliefs, that is, 
it goes against what is commonly believed or accepted.
In the case of the example in Table~\ref{tab:examples-symcon},
the contents of two sentences, \textit{Some humans are not living things} and \textit{None of the animals are human} are contrary to common sense.

There are 95 instances of \textit{symbolic},
167 instances of \textit{consistent},
and 102 instances of \textit{inconsistent}.
For cases where the judgment of belief consistency is unclear, we classify them as \textit{others}. We encountered 11 instances that fell into this category.

\begin{table*}[ht!]
\centering
\begin{tabular}{l | l | l} \hline
\textbf{Type of syllogisms} & \textbf{Language} & \textbf{Example} \\ \hline
\ShowExp{\textbf{AAA}}
{All B are A.}
{All B are C.}
{All A are C.}
{すべての \ B \ は \ A \ である}
{すべての \ B \ は \ C \ である}
{すべての \ A \ は \ C \ である}
\\ \hline\hline
\ShowExp{\textbf{AOO}}
{All chimpanzees are animals.}
{Some animals are not primates.}
{Some primates are not chimpanzees.}
{すべてのチンパンジーは動物である。}
{ある動物は霊長類でない。}
{ある霊長類はチンパンジーでない。}
\\ \hline\hline
\ShowExp{\textbf{OAO}}
{Some ghosts are not students.}
{All students are humans.}
{Some humans are not ghosts.}
{ある幽霊は生徒でない。}
{すべての生徒は人間である。}
{ある人間は幽霊でない。}
\\ \hline\hline
\ShowExp{\textbf{EAO}}
{No robot is human.}
{Every human being is a living organism.}
{A certain living organism is not a robot.}
{どのロボットも人間でない。}
{すべての人間は生物である。}
{生物のあるものはロボットでない。}
\\ \hline\hline
\ShowExp{\textbf{AII}}
{All humans are animals.}
{Some robots are animals.}
{Some humans are robots.}
{すべての人間は動物である。}
{あるロボットは動物である。}
{ある人間はロボットである。}
\\ \hline
\end{tabular}
\caption{Examples of syllogistic inference problems
where conversion errors give wrong answer.
The correct label for all the examples is \textit{neutral}.
The English sentences (\textbf{P1}, \textbf{P2}, \textbf{C})
in each example correspond to the respective Japanese sentences.
}
\label{tab:converr}
\end{table*}

\section{Human-like biases}
\label{sec:bias}

Based on the above classification of the types of syllogistic inferences and sentences, we examine three types of human-like biases that can cause reasoning errors:
belief biases, conversion errors, and atmosphere effects.
We annotated information to each inference in the dataset to make explicit which inferences are misjudged by these biases.

\subsection{Belief Biases}
\label{sec:belief}

Belief bias is one of the most well-known biases causing inference errors and has been applied to various types of logical inferences including syllogisms and Wason's selectional task~\cite{evans1989bias,newstead1992source,evans1993human}.
It is widely recognized that people tend to have trouble in determining
whether an inference is valid when it includes a sentence contrary to common sense.
For example, the inference that is labelled as \textit{inconsistent} in Table~\ref{tab:examples-symcon},
repeated here, has inconsistent sentences \textbf{P2} and \textbf{C}:

\begin{center}
\syllogism
{Some animals are human beings.}
{All animals are tomatoes.}
{Some humans are tomatoes.}
\end{center}

\noindent
Although the correct label for this problem is \textit{entailment}, the fact that the conclusion \textbf{C} is contrary to beliefs may lead some to judge it as \textit{contradiction} instead of \textit{entailment},
regardless of its logical validity.
Similarly, in the following example, while \textbf{P2} is contrary to our beliefs, the conclusion \textbf{C} remains consistent. Hence, one might judge the inference as \textit{entailment} rather than \textit{neutral} due to the belief-consistency of the conclusion.
\begin{center}
\syllogism{All canines are animals.}
{All animals are robots.}
{No canine is a robot.}   
\end{center}

As mentioned in Section~\ref{sssec:belief},
when either one of the premises or the conclusion
is inconsistent with our beliefs,
we assigned the \textit{inconsistent} label to the inference.
We investigate whether or not NLI models are
influenced by this type of belief biases.

\begin{table*}[ht!]
\centering
\begin{tabular}{l | l | l} \hline
\textbf{Type of syllogisms} & \textbf{Language} & \textbf{Example} \\ \hline
\ShowExp{\textbf{AOE}}
{All animals are living things. }
{Some humans are not living things.}
{None of the animals are human.}
{すべての動物は生物である。}
{ある人間は生物でない。}
{どの動物も人間でない。}
\\ \hline\hline
\ShowExp{\textbf{OAI}}
{A certain police officer is not a public servant. }
{All human beings are public servants.}
{Some police officer is a human being.}
{ある警察官は公務員でない。}
{すべての人間は公務員である。}
{ある警察官は人間である。}
{}
{}
{}
\\ \hline
\end{tabular}
\caption{Examples of syllogistic inference problems
where atmosphere effects can give wrong answer.
The correct label for all the examples is \textit{neutral}.
The English sentences (\textbf{P1}, \textbf{P2}, \textbf{C})
in each example correspond to the respective Japanese sentences.
}
\label{tab:atmos}
\end{table*}

\subsection{Conversion Errors}
\label{ssec:conversion}

Conversion errors are errors in syllogisms caused by the incorrect interpretation of terms that appear in premises.
There are at least two types of errors, called \textit{illicit conversion}~\cite{Wilkins1928,newstead1989interpretational,Geurts2003-GEURWQ}:
\begin{enumerate}
\item The tendency to interpret \textit{All A are B} as equivalent to \textit{All B are A} (\textbf{A}-type)
\item The tendency to interpret \textit{Some A are not B} as equivalent to \textit{Some B are not A} (\textbf{O}-type).
\end{enumerate}
Note that \textit{All A are B} and \textit{Some A are not B}
mean
$A \subseteq B$ and $A \cap \overline{B} \neq \emptyset$,
respectively, in the standard predicate logic\footnote{
In traditional syllogisms, \textbf{A}-type sentence
\textit{All A are B} implies that $A$ is not empty.
The BAROCO dataset follows this traditional interpretation
(the \textit{existential import} of universal expressions)
when annotating the gold labels.
},
hence terms $A$ and $B$ are not convertible.
Table~\ref{tab:converr} shows
some examples of syllogistic inference problems
where conversion errors cause wrong answer.

We identified the syllogisms in which the correct answer is \textit{neutral} and whose premises contain a sentence of the form \textit{All A are B} or \textit{Some A are not B}, and whose correct answer changes from \textit{neutral} to \textit{entailment} by applying conversion to either one or both premises.
We annotate the \textit{conversion} label to this type of problems.
In the original dataset, there are only 10 such problems.
Thus we expanded the dataset by adding more \textit{conversion} problems
that have the types not included in the original dataset.
We fixed a set of schematic types to be added and 
obtained instances of these types
by substituting abstract terms with concrete nouns in the dataset.
In total, there are 70 problems to which the \textit{conversion} label is assigned in the NeuBAROCO dataset.

\subsection{Atmosphere Effects}
\label{ssec:atmos}
Atmosphere effects are one of the inferential biases that can be traced back to studies in the 1930s~\cite{woodworth1935atmosphere,khemlani2012theories}.
It can be interpreted as
two principles~\cite{chater1999probability}:
\begin{enumerate}
\item \textit{The principle of quality}: if one or both premises
are negative (\textbf{E}-type or \textbf{O}-type),
the conclusion should be negative;
otherwise, it is positive (\textbf{A}-type or \textbf{I}-type).
\item \textit{The principle of quantity}:
if one or both premises are particular (\textbf{I}-type or \textbf{O}-type),
then the conclusion will be particular; otherwise, it is universal
(\textbf{A}-type or \textbf{E}-type).
\end{enumerate}
Previous psychological experiments based on the original BAROCO data (for Japanese) have shown that \textbf{O}-type inference is particularly difficult for logically untrained human participants~\cite{SHIKISHIMA2009256}.
Thus among various patterns,
we focus on the cases where at least one premise is an \textbf{O}-type or \textbf{I}-type sentence.
We assign the \textit{atmosphere} label to an inference
if its correct answer is \textit{neutral} and 
(1) at least one of the premises is \textit{O}-type and the conclusion is either
\textbf{E}-type, \textbf{I}-type, or \textbf{O}-type or (2)
at least one of the premises is \textbf{I}-type and the conclusion
is either \textbf{I}-type or \textbf{O}-type.

Table~\ref{tab:atmos} shows some representative examples of syllogisms satisfying these conditions.
In total, there are 104 problems labelled as \textit{atmosphere} in the dataset.

\section{Experiments}
\label{sec:experiment}

\subsection{Experimental settings}
\label{ssec:setting}

We evaluate syllogistic reasoning ability of deep neural networks, and in particular state-of-the-art large language models
using our NeuBAROCO dataset.
We evaluate transformer-based pre-trained language models, RoBERTa~\cite{liu2019roberta} and BART~\cite{lewis-etal-2020-bart}, both being fine-tuned with the MultiNLI dataset~\cite{williams-etal-2018-broad}.
We use the models available in the \texttt{transformers} library,
\texttt{roberta-large-mnli} and \texttt{facebook/bart-large-mnli}.\footnote{\url{https://github.com/huggingface/transformers}}
We also evaluate OpenAI's GPT-3.5 model, an improved version of GPT-3~\cite{brown2020language}.
We use OpenAI's ChatGPT API with the \texttt{GPT-3.5-turbo} model.\footnote{
\url{https://platform.openai.com/docs/model-index-for-researchers}}

To test whether the models show belief biases, conversion errors, and atmosphere effects, we tested how well the models can answer correctly to the problems labelled as \textit{inconsistent}, \textit{conversion}, and \textit{atmosphere},
and compared the accuracies with the total average on the NeuBAROCO dataset.

\begin{table*}[ht!]
\centering
\begin{tabular}{l|l|c| ccc} \hline
\textbf{Language} & \textbf{Models} & \textbf{All} & \textbf{Entailment} & \textbf{Contradiction} & \textbf{Neutral}\\ \hline
\multirow{3}{*}{English} & RoBERTa & 34.67  & 62.30 & 74.65 & 0.55\\ \cline{2-6}
& BART & 35.20 & 55.74 & 83.10 & 2.75 \\ \cline{2-6}
& GPT-3.5 & 51.73 & 79.51 & 38.03 & 38.46\\ \hline
Japanese & GPT-3.5 & 48.27 & 80.33 & 54.93 & 24.18 \\ \hline
\end{tabular}
\caption{Accuracy (\%) of the models on 
all the inference problems and each correct inference label.
}
\label{tab:res}
\end{table*}

\begin{table*}[ht!]
\centering
\begin{tabular}{l|l| cc} \hline
\textbf{Language} & \textbf{Models} & \textbf{Basic} & \textbf{Extended} \\ \hline
\multirow{3}{*}{English} & RoBERTa & 31.13 & 54.39 \\ \cline{2-4}
& BART & 31.13 & 57.89 \\ \cline{2-4}
& GPT-3.5 & 51.57 & 52.63 \\ \hline
Japanese & GPT-3.5 & 46.86 & 56.14 \\ \hline
\end{tabular}
\caption{Accuracy (\%) of the models on 
basic and extended syllogisms.
}
\label{tab:ressyl}
\end{table*}

\begin{table*}[ht!]
\centering
\scalebox{0.95}{
\begin{tabular}{l|l|l| ccc | cc} \hline
\textbf{Language} & \textbf{Models} & \textbf{All} & \textbf{Symbol} & \textbf{Consistent} & \textbf{Inconsistent} & \textbf{Conversion} & \textbf{Atmosphere} \\ \hline
\multirow{3}{*}{English} & RoBERTa & 34.67 & 24.21 & 46.11 & 22.55 & 0.0 & 0.0\\ \cline{2-8}
& BART & 35.20 & 34.74 & 45.51 & 15.69 & 1.43 & 0.96 \\ \cline{2-8}
& GPT-3.5 & 51.73 & 61.05 & 56.89 & 31.37 & 25.71 & 39.42 \\ \hline
Japanese & GPT-3.5 & 48.27 & 55.79 & 56.29 & 25.49 & 21.43 & 22.12 \\ \hline
\end{tabular}
}
\caption{Accuracy (\%) of the models on
each type of syllogistic inferences
and biases.
}
\label{tab:resbias}
\end{table*}

\subsection{Results and discussion}
\label{ssec:results}

\subsubsection{Overall results}
\label{sssec:overall}

Table~\ref{tab:res} shows the overall accuracy of each model
and the accuracy on each correct label.
Table~\ref{tab:ressyl} shows the accuracy of the models
on basic and extended syllogisms.

\paragraph{RoBERTa}
The overall accuracy
was low (34.67\%). The accuracy for the \textit{contradiction} problems was very high (74.65\%), although the accuracy of the \textit{neutral} problems was very low (0.55\%). The accuracy of the \textit{entailment} problems was between the two (62.3\%).
The accuracy on extended syllogism was higher than basic syllogism (54.39\% and 31.13\%).

\paragraph{BART}
The results on BART shows the same tendency.
The overall accuracy was low (35.2\%). BART answered correctly in most \textit{contradiction} cases, but less in \textit{entailment} and still less in \textit{neutral} (83.1\%, 55.74\% and 2.75\%). The performance on extended syllogism was better than that on basic syllogism (57.89\% and 31.13\%).

\paragraph{GPT-3.5}
The overall accuracy was 51.73\%.
The accuracy on the \textit{entailment} problems was very high (79.51\%), while that on the \textit{contradiction} and \textit{neutral} problems
 were low (38.03\% and 38.46\%).
We found little difference between basic and extended syllogism (51.57\% and 52.63\%).

\subsubsection{Results on problems concerning biases}
\label{sssec:resbias}

Table~\ref{tab:resbias} shows the accuracy of the models
on each type of syllogistic inferences and biases.

\paragraph{Belief Biases}
Among the three types of inferences, \textit{symbol},
\textit{consistent}, and \textit{inconsistent},
the performance on the \textit{inconsistent} cases was the lowest in every model. We found a significant difference among the models on which of the \textit{symbol} or \textit{consistent} cases they answered most accurately. 
GPT-3.5 performed better in \textit{symbol} and \textit{consistent} than \textit{inconsistent} (61.05\%, 56.89\% and 31.37\%). For RoBERTa, the percentage of correct answers to the \textit{consistent} cases was higher than in \textit{inconsistent} and \textit{symbol} cases (46.11\%, 22.55\% and 24.21\%). The percentages of correct responses of BART were, in order of highest to lowest, \textit{consistent}, \textit{symbol}, and \textit{inconsistent} (45.51\%, 34.74\% and 15.69\%).

Overall, the results show a tendency that GPT-3.5 outperforms both RoBERTa and BART models in symbolic reasoning.
Also, GPT-3.5 performed equally well on \textit{symbol} and 
\textit{consistent} problems. This suggests that \textit{symbol} and 
\textit{consistent} are relatively easy to handle in that both types of problems are not contrary to beliefs.
These results contrast with the results on RoBERTa, which performed almost equally on \textit{symbol} and \textit{inconsistent} problems.

\paragraph{Conversion Error} 
Regarding the \textit{conversion} problems,
all models exhibit low performance.
A striking difference exists between RoBERTa and BART, on one hand, and GPT-3.5, on the other.
While RoBERTa and BART hardly answered correctly (0\% and 1.43\%), GPT-3.5 performed better, answering correctly almost a quarter of the cases, which is about half of the overall average (25.71\%). 
The results show that all the models performed poorly on 
the problems where incorrect responses are made by conversion errors.

\paragraph{Atmosphere Effects}
We found that the performances of the models were
notably lower in \textit{atmosphere} cases. 
While RoBERTa and BART hardly provide correct answer to \textit{atmosphere} cases (0\% and 0.96\%),
GPT-3.5 performed better (39.42\%).

\subsubsection{Results on the Japanese GPT model}

The results on the Japanese GPT model
shows the strikingly same tendency as the English GPT models.
One notable exception is the performance on 
the \textit{contradiction} problems (see Table~\ref{tab:res}),
where the Japanese GPT-3.5 model performed better than
the English GPT-3.5 models.
By contrast, the performance on \textit{atmosphere} problems
was worse than that of the English model.

\section{Conclusion}

In this paper, we investigated syllogistic reasoning ability of current large language models in focusing on human-like biases that have been studied in the context of cognitive science of human reasoning. The experiments indicated that the state-of-the-art models fail for problems where errors are caused by various human-like biases, and that there is large room for improvement in deductive reasoning capabilities of large language model.

Among other things, our results on conversion errors
suggest the importance of distinguishing the problems of interpreting sentences (in particular, interpreting quantifiers and negation) from
the problem of performing logical inferences.
For conversion cases, there is abundant room for discussion on at which level the models mistook. Further inquiry into this issue could provide insight into a better understanding of the behavior of neural models.

There remain many issues to be addressed.
First, although we tested the models with no solved examples, experiments on few-shot learning will be insightful. \citet{dasgupta2022language} reported that the models performed syllogistic reasoning better with few-shot learning, whether or not the content of the inferences were consistent with common-sense beliefs. In addition, we suppose that the performances can be different depending on the wording in instructions. Further research will contribute to improvement of instructions.

Second, we showed that the models performed worse in the cases in which humans have troubles because of some well-known human biases.
It is left for future work to make more detailed comparisons
between humans and neutral models, which is a promising research direction
since the original BAROCO data on human reasoning ability is available.

Finally, it is interesting to consider
extended forms of natural logic inferences other than basic syllogisms,
including relational syllogism with transitive verbs and comparatives
and those inferences with generalized quantifiers such as \textit{most}.
It is left for future work to examine whether the models show similar biases
for such extended syllogism inferences.

\newpage

\section*{Author Contributions}

The three authors, Risako Ando, Takanobu Morishita, and Hirohiko Abe, made equal contributions as core authors.

\section*{Acknowledgements}
We express our gratitude to the anonymous reviewers for their helpful feedback, which has contributed to the improvement of this work.
We would like to express our thanks to Prof.\,Chizuru Shikishima for the discussion on earlier versions of the BAROCO inference tasks collection.
This work is partially supported by JST, CREST grant number JPMJCR2114, MEXT-JSPS Kakenhi MKK279H and MKK477J.

\bibliography{custom}
\bibliographystyle{acl_natbib}

\end{document}